\documentclass[journal]{IEEEtran}
\usepackage[utf8]{inputenc}
\usepackage[T1]{fontenc}
\usepackage{textcomp}

\usepackage{graphicx}
\usepackage{hyperref}
\usepackage{amsmath}
\usepackage{amsfonts}
\usepackage{siunitx}
\usepackage{longtable,tabularx}
\usepackage{booktabs}
\usepackage{multirow}
\usepackage{subfig}
\usepackage{xcolor}
\usepackage{lineno}
\usepackage{ntheorem}
\theoremseparator{:}
\newtheorem{hypothesis}{Hypothesis}

\title{\LARGE \bf Predictive Probability Density Mapping for Search and Rescue Using An Agent-Based Approach with Sparse Data}

\author{
	{Jan-Hendrik Ewers$^{1}$, David Anderson$^{2}$, and Douglas Thomson$^{2}$}
	\thanks{This work has been submitted to the IEEE for possible publication. Copyright may be transferred without notice, after which this version may no longer be accessible.}
  \thanks{This work was supported by the Engineering and Physical Sciences Research Council, Grant/Award Number: EP/T517896/1-312561-05}
	\thanks{$^{1}$ Aerospace Sciences Research Division, University of Glasgow, Scotland 
	{\tt \small j.ewers.1@research.gla.ac.uk
	}} 
	\thanks{$^{2}$ Aerospace Sciences Research Division, University of Glasgow, Scotland 
	{\tt \small \{dave.anderson, douglas.thomson\}@glasgow.ac.uk}
}}

\renewcommand{\vec}[1]{\mathbf{#1}}

\usepackage{nomencl}
\newcommand{\nomdef}[3]{#1 (#2)\nomenclature[#3]{#2}{#1}}
\newcommand{\acrdef}[2]{\nomdef{#1}{#2}{A}}

\makenomenclature

\usepackage{etoolbox}
\renewcommand\nomgroup[1]{%
  \item[\bfseries
  \ifstrequal{#1}{A}{Acronyms}{%
  \ifstrequal{#1}{S}{Symbols}{}}%
]}

\DeclareSIUnit\km{\kilo\metre}
\DeclareSIUnit\kmsq{\km\squared}

\newcommand{\tech}[1]{\emph{#1}}

\newcommand*\mean[1]{\bar{#1}}

\newcommand{\citet}[1]{\cite{#1}}

\newcommand{\Hikersolo}{\emph{Hiker (solo)}~}
\newcommand{\reffig}[1]{Fig.~\ref{#1}}
\newcommand{\refeqn}[1]{Eqn.~\ref{#1}}
\newcommand{\reftbl}[1]{table~\ref{#1}}
\newcommand{\refsect}[1]{Sect.~\ref{#1}}
\newcommand{\refhyp}[1]{Hyp.~\ref{#1}}
\usepackage{xcolor}

\usepackage{xspace}

\newcommand\rawheadto[2]{#1ead2#2\xspace}
\newcommand\Headto[1]{\rawheadto{H}{#1}}

\newcommand\headtowater{\Headto{Water}}
\newcommand\headtopaths{\Headto{Paths}}
\newcommand\headtobuildings{\Headto{Buildings}}
\newcommand\headtotrees{\Headto{Trees}}

\graphicspath{{sections/conclusion/figures/}{sections/intro/figures/}{sections/method/figures/}{sections/results/figures/}{sections/DOE/figures/}}

\begin{document}

\maketitle

\begin{abstract}						
	Predicting the location where a lost person could be found is crucial for
	search and rescue operations with limited resources.
	To improve the
	precision and efficiency of these predictions,
	simulated agents can be
	created to emulate the behavior of the lost person.
	Within this study, 
	we introduce an innovative agent-based model designed
	to replicate diverse psychological profiles of lost persons, 
	allowing these agents to navigate real-world landscapes while making
	decisions autonomously 
	without the need for location-specific training. 
	The probability distribution map depicting the potential location of the lost person
	emerges through a combination of Monte Carlo simulations and
	mobility-time-based sampling. 
	Validation of the model is achieved using real-world Search and Rescue data to train a Gaussian Process model. 
	This allows generalization of the data to sample initial starting points for the agents during validation.
	Comparative analysis with historical data
	showcases promising outcomes relative to alternative methods. 
	This work introduces a flexible agent that can be employed in search and rescue operations, 
	offering adaptability across various geographical locations.
\end{abstract}


\section{Introduction}
\label{sect:intro}

\acrdef{Search and Rescue}{SAR} of vulnerable people is unfortunately a common task for the Police and other emergency services.
Organizations like the Centre for Search and Rescue\cite{perkins_missing_2011} and the Grampian Police\cite{grampianpolice_missing_2007} carry out research and training in areas related to SAR and while their training and published papers offer a valuable resource to the people responsible for finding a
\acrdef{Lost Person}{LP},
they focus only on land-based search i.e.
directing teams of individuals.
This is a slow and methodical process that would undoubtedly benefit from the assistance of an airborne surveillance platform.
As a result of rapid advances in the drone sector over the last decade,
multirotors capable of carrying high-powered sensor payloads have become cheaper and more accessible than ever before.
Consequently,
several concept evaluation trials for measuring the efficacy of incorporating drones into the search process have recently been undertaken in Scotland\cite{skeleton_remotely_2020}.

Air-based SAR is a core operational requirement of the \acrdef{Police Scotland Air Support Unit}{PSASU}\cite{skeleton_remotely_2020} and an up-and-coming tool for \acrdef{Scottish Mountain Rescue}{SMR}\cite{carrell_flying_2022}.
Both PSASU and SMR traditionally use large helicopters which have the same limitations: cost,
transit time,
availability,
and weather.
The latter is a common issue during SMR rescues as near-perfect conditions are required for safe helicopter evacuation.
Similarly,
PSASU has their only Eurocopter EC135 based in Glasgow,
Scotland.
If this helicopter was required in remote locations,
such as the Isle of Orkney,
this life-saving resource would be multiple hours away from full deployment.
Hence,
PSASU and SMR are placing a small fleet of drones in key locations around Scotland for rapid deployment to assist in LP search operations where seconds matter.
These will not be replacing the helicopter,
but are rather intended to complement its operations.

Whilst rapid deployment is a key requirement for the usage of \acrdef{Unmanned Aerial Vehicle}{UAV} in an LP search,
the cost of their usage must also be explored.
During a rescue operation,
cost can be defined in terms of time,
personnel,
and money.
The latter is calculated pre- or post-mission,
but the time and personnel cost are what the search leader balances.
Having more personnel available
means that more searchers can actively cover the search area and this could directly lead to faster find time.
However,
personnel are not an infinite resource and people need to be effectively assigned.
Therefore,
computational assistance to the UAV-based search segment is a key element in freeing up resources.

The mission profile for UAVs in a SAR setting is solely for search. 
This requires finding the LP as quickly as possible and relaying this information back to the search leader so that they can organize the rescue operations.
The search further breaks down into prediction,
flying,
and sensing.
Flying and sensing are done in real-time\cite{brown_trajectory_2020},
but the prediction of where the LP could be can be done en route to the scene.
By creating a \acrdef{Probability Distribution Map}{PDM},
the search leader and their available resources can be informed on where the most \emph{likely} locations are to find the LP.
This,
in theory,
results in a decrease in the time to find an LP that could save lives.

One key piece of information in the prediction process is using data to inform and validate the models.
Historical data collected from SAR cases agrees that significant behavioral profiles exist\cite{koester_lost_2008,
perkins_missing_2011}.
This means that a solo hiker behaves differently from an elderly person with dementia when moving across a landscape,
which ultimately affects the location found.
By using this a priori information along with other location data,
these PDMs can be highly customized on a per-location and per-person basis resulting in better PDMs\cite{ewers_optimal_2023}.

A crucial constraint is the lack of available data. Even the largest SAR database ISRID\cite{koester_international_2000} has only $50,000$ data points, with many not including geospatial information. PDMs need large amounts of data that cannot be gathered effectively through simple data logging by various bodies. This drives the need for models to generate the data so that the likes of SMR and PSASU always have readily available PDMs.

This research builds upon prior work in the field\cite{ewers_gis_2023}, where limitations due to high errors of over $20\%$ were identified for three out of the four core behavioral models (introduced in \refsect{sect:lp_behavior}) during validation. To address these shortcomings, we propose extensions to these specific aspects of the model naming this iteration J2.
Furthermore, this work introduces a supervised machine-learning approach to the model validation stage by up-sampling sparse historical data. By leveraging real-world data, the new proposed approach allows for a more robust assessment of the models' effectiveness, potentially leading to significant improvements in real-world applications.

PDM generation algorithms of merit are discussed in \refsect{sect:lit_review},
followed by the methodology used in this research in \refsect{sect:method}.
The design of experiment is introduced in \refsect{sect:doe}.
Results are presented and discussed in \refsect{sect:results_and_discussion}.
Finally,
\refsect{sect:conclusion} concludes this paper and discusses future work.

\section{Current PDM Generation Algorithms}
\label{sect:lit_review}

The most basic form of PDM generation is the \acrdef{Euclidean Distance
	Circle}{EDC} as described by
\citet{koester_lost_2008} and \citet{heth_characteristics_1998}. This method works on bounding a search area within a
circle based on the straight line distance from statistics about how far away
from the \acrdef{Place Last Seen}{PLS} an LP might be. 
Whilst statistically sound, 
this method fails to incorporate the terrain of the search area at hand. 
Approximately 75\% of wilderness SAR incidents happen in mountainous regions\cite{sava_evaluating_2016} where there are many barriers to travel around. 
\citet{heth_characteristics_1998}
outlines a modified EDC algorithm that allows incorporating impassable features into the resultant PDM, 
however this is only possible through manual intervention by the searcher. 
This makes it unsuitable for any automatic PDM generation.
As well as the inability to incorporate geographical features, the search area
size radically changes depending on the LP profile. From data seen in
\citet{koester_lost_2008}, the $95\%$ radius for \tech{Child 12-15} is $5.65$
times larger than the \tech{Child 10-12} category. This makes it hard to focus
resources in a SAR scenario based purely on the EDC.

Like the EDC model, the Watershed model from \citet{sava_evaluating_2016} and
\citet{doke_analysis_2012} relies on historical distance data to be constructed.
\citet{doke_analysis_2012} found that in Yosemite National Park, 48\% of LPs were
found within the original (0th) watershed (an area of land that separates water
flowing into different areas), and 38\% were found in the adjacent watersheds
with \citet{sava_evaluating_2016} finding similar results. 
However,
whilst it
incorporates information about the terrain,
it is limited to a single source of data.
Using more models, or data, is beneficial in creating more nuanced models, as
discussed in \citet{surowiecki_wisdom_2014}. This also gives the benefit of
generating more customized PDMs for a single search area. A benefit of using
simpler models is that less computational, or human effort, is required to
generate a PDM. A searcher in the hills may not have access to a powerful
computer to calculate specific PDMs or the time to wait for results.

Mobility models, such as by \citet{doherty_analysis_2014}, try to estimate how
far the LP may have traveled in all directions from the PLS. This is done by
combining simulations of walking speed with environmental effects that may
affect their motion. For example, walking through a dense forest will be slower
than over a grassy field. Another approach is considering what the path of least
resistance is to a given coordinate from the PLS. 
This can be done by calculating the cost of passing through a subset of the area and  applying a path-planning algorithm like Randomly Exploring Random Tree
\cite{lavalle_rapidlyexploring_1998}.
Other models may use Accumulated Cost Surface
\cite{douglas_leastcost_1994}
algorithms as they tend to be included in 
\acrdef{Geographic Information System}{GIS}
software.

An advancement on the mobility model is the 
\acrdef{Travel Time Cost Surface Model}{TTCSM}\cite{brentfrakes_national_2015}.
This model uses the notion
of \tech{percent of maximum travel speed} to evaluate the time taken to
travel through a cell. This value is assigned by a location expert and can vary
for different search areas. Limits are assigned to the area through this concept
by giving it a value of $0$, such as for in a lake or a
cliff 
(defined as any slope steeper than $31\si\degree$).
A major factor affecting the walking speed of the LP is the slope angle, which
is extrapolated from the digital elevation map. TTCSM uses the
Tobler model\cite{tobler_three_1993} for walking speed which provides a continuous
estimation based on gradient $m = \tan(s)$, where $s$ is the slope angle.

\citet{drexel_network_2018} developed the Travel Time Network Model which builds on TTCSM and focuses on trail networks. 
In this model, 
network theory is used rather than the rasterized method from before. 
Every edge 
(a path)
is given a cost associated with the time taken to traverse.
The network is then used to generate the PDM using a
\tech{service area analysis} 
at defined threshold values. 
These threshold values are the mobility times from 
\citet{koester_lost_2008}.
An evident drawback of this approach is the assumption that the PLS is on a trail,
and that there are sufficient trails for the model to work effectively. 
The latter is particularly important as this model would not work as well in the rural Scottish Highlands as it would in the English Lake District.
However, the approach is viable when these assumptions have been met.

\citet{hashimoto_agentbased_2019}
and
\citet{heintzman_anticipatory_2021}
developed a six-behavior algorithm to simulate an agent navigating a 2D grid.
Each time step involves selecting a behavior through a learned weighting vector.
This chosen behavior assigns probabilities to the agent's current cell and its eight surrounding cells.
For example, \tech{staying put} would set the surrounding cells to $0$ and the current cell to $1$, whereas \tech{random walk} sets all cells to $\frac{1}{9}$.
The agent subsequently picks one of the nine possible cells based on these probabilities and moves to it.
This iterative process is then repeated.
The weighting vector is constructed by evaluating all permutations 
(in steps of $\frac{1}{6}$)
and comparing the model data against the statistics from \citet{koester_lost_2008}.
This was then further developed in
\citet{hashimoto_agentbased_2022}
using a leave-one-out analysis to further improve the accuracy of the resultant behavior weighting vector compared to the real-world data from
\citet{koester_lost_2008}. 
However, 
this vector has to be learned for every new location as results differed substantially between trials.

Thus, J2 (like J1) combines the combine the behavior-based approach from
\citet{hashimoto_agentbased_2022}
and a modification of the network-based approach from 
\citet{drexel_network_2018} to meet its assumptions,
 to further use the physical information about the landscape. 
Unlike \citet{hashimoto_agentbased_2022}, however, behaviors are segmented by final find location rather than by movement types, which aims to prevent the per-location training.

\section{Methods}
\label{sect:method}

The method outlined in this section, 
named J2,
has two parts. In \refsect{sect:method_path_generation}, the
Monte Carlo path generation using the LP surrogate is introduced. The sampling of the paths to generate locations found is then developed in \refsect{sect:method_sampling}.

The data relating to lost person behavior is primarily sourced from
\citet{perkins_missing_2011} with missing data being taken from
\citet{koester_lost_2008}. Both sources consider multiple categories of LP, such
as \tech{Child 10-12} or \tech{Climber}. 
However, one of the largest datasets reported by both is 
\Hikersolo. All parameters used in this research are based on this category from here on in.

\subsection{Path Generation}
\label{sect:method_path_generation}

At its core, 
every simulation run is a behavior traversing the landscape from a starting point until the path length goes above
$D_{\max}$. 
For this stage, the agent does not tire. 
Whilst this is an unrealistic assumption by itself, 
the LP movement times are introduced during the sampling stage defined in \refsect{sect:method_sampling}.

The LP is modeled as an agent navigating a 2D grid with square
cells of shape $5\si\metre \times 5\si\metre$. Its viewpoint sits at a constant
$1.6\si\metre$ above the surface, allowing it to see over smaller obstacles like
rocks or long grass. 
At each time step,
the agent moves to the next cell and the total distance increases accordingly until it reaches the termination distance 
$D_{\max}$. The simulation will then terminate at the next distance check.

The starting points are then sampled from the bivariate Gaussian probability distribution function $PDF_\textit{start}(\vec x)$ which is defined as
\begin{gather}
	PDF_\textit{start}(\vec x) = \frac{1}{\sqrt{4\pi^2\det{\vec\sigma}}} \exp{\left[-\frac{1}{2}(\vec x - \vec \mu )^T\vec\sigma^{-1}(\vec x - \vec \mu) \right]}
	\label{eqn:gaussian}\\
	\vec \sigma
	=
	\begin{bmatrix}
		\sigma_{xx} & \sigma_{xy} \\
		\sigma_{yx} & \sigma_ {yy} 
	\end{bmatrix} \\
	\vec \mu
	=
	\begin{bmatrix}
		\mu_x \\
		\mu_y
	\end{bmatrix} 
\end{gather}
centered around a given start location $\vec \mu$. 
In a real SAR mission, this would be the PLS. 
The purpose of $PDF_\textit{start}$ is to model the uncertainty, through 
$\vec \sigma$,
 of the reported PLS. 
SMR reports a PLS using the 6-digit Ordnance Survey National Grid reference, 
which reduces the British Isles into $100 \times 100 \si{\m}$ cells. A PLS described by this system has a $\pm 100 \si\m$ error in $x$ and $y$, with a total magnitude of the error being $\pm 141.42 \si{\m}$. 
This would result in a variance of $\sigma_{xx} = \sigma_{yy} = 10,000$.
To validate this model, \refsect{sect:doe} outlines how $\vec \mu$ is generated based on real-world data.

\subsubsection{Lost Person Behavior}
\label{sect:lp_behavior}

\begin{table}[tb]
	\centering														
	\caption{Land cover category location found data for a \Hikersolo from \citet{perkins_missing_2011} and the associated behaviors}
	\label{tbl:perkins_solo_hiker}
	\begin{tabular}{@{}lrrr@{}}
		\toprule
		Land cover category & n  & \%  & Behavior \\\midrule
		Open Ground    & 53 & 40.8 & n/a \\ 
		Travel Aid     & 33 & 25.4 & \headtopaths \\
		Building       & 30 & 23.1 & \headtobuildings \\
		Linear Feature & 9  & 6.9  & \headtopaths \\
		Trees          & 4  & 3.1  & \headtotrees \\
		Water          & 1  & 0.8  & \headtowater \\ \bottomrule
	\end{tabular}
\end{table}

To simulate the behavior of an LP, four behaviors were created. Each behavior corresponds to a land cover category that the agent will attempt to travel to. Throughout the simulation, the agent keeps the same behavior. The intent of this is to emulate an LP over a large number of data points. From the data shown in \reftbl{tbl:perkins_solo_hiker}, a behavior can be selected $P\%$ of the time. \tech{building}, \tech{trees} and \tech{water} were given individual behaviors, but \tech{linear feature} and \tech{travel aid} were merged under \headtopaths. 
This was done because \tech{linear feature} is defined as being a \tech{stream/ditch} or \tech{wall/fenced line}, and these features often follow paths and roads. As well as this, a \tech{stream} is already being handled by the \tech{water} behavior (as further outlined below).
Finally, \refhyp{hyp:open_ground_find_location} (with later empirical evaluation) was employed to generalize the \tech{open-ground} find data.

\begin{hypothesis}
	The open-ground land cover category location found data naturally results from an LP trying to navigate to the other possible locations.
	\label{hyp:open_ground_find_location}
\end{hypothesis}

\begin{table*}[htb]
	\centering
	\caption{GIS maps used in this research}
	\label{tbl:list_of_gis_maps}
	\begin{tabular}{@{}llll@{}}
		\toprule
		Map                       & Description                                                                 & Source                                    \\ \midrule
		Digital Elevation         & \acrdef{Light Detection and Ranging}{LiDAR} composite Digital Terrain Model & \citet{ascgeospatialdata_lidar_2014}       \\
		Land Cover ID             & Land cover ID map                                                     & \citet{geopackagegeospatialdata_land_2021} \\
		Cumulative Catchment Area & The area draining through a point                                           & \citet{morris_digital_1990}                \\
		Water Surface Type        & Most hydrologically significant surface type                                & \citet{morris_digital_1990}                \\
		Water Outflow Direction   & Shows the direction of the overland flow                                    & \citet{morris_digital_1990}                \\
		Road Network              & Road and path network                                                       & \citet{ordnancesurvey_os_2021}             \\ \bottomrule
	\end{tabular}
\end{table*}

The GIS maps utilized in this study are presented in \reftbl{tbl:list_of_gis_maps}. It is crucial to clarify that, although land cover ID and land cover category may seem similar, they differ. The former is exclusively employed within the viewshed-based behaviors, as outlined in \refsect{sect:vector_based_behavior}, and is sourced from \citet{geopackagegeospatialdata_land_2021}. On the other hand, the latter influences the meta-behavior of the LP and originates from \citet{koester_lost_2008} and \citet{perkins_missing_2011}.

\paragraph*{\headtowater}
\label{sect:head2water}

The methodology for the \headtowater behavior was left unchanged from \citet{ewers_gis_2023} with the usage of a vector field following model. 
The GIS data outlined in
\reftbl{tbl:list_of_gis_maps}
shows that three of the six maps (cumulative catchment area, water surface type, and water outflow direction) contain hydrological information, 
with the water outflow direction map being used as a direct input to the vector field follower. 
To increase the fidelity of the model,
the cumulative catchment area and the water surface type can further be used to force the agent to navigate around large enough bodies of water. 
This was done by evaluating the position of the agent in the next time step, 
and if the water surface type at the future position was either 
\tech{lake}, 
\tech{sea},
or 
\tech{river} then a scaled probability is calculated of carrying out this step.
This is done through
\begin{equation}
	p(a) = 1-\begin{cases}
	1,&a\geq b \\
	\frac{a}{b},&\text{else}
	\end{cases}
\end{equation}
where
$v \in [0,\infty)$
is the cumulative catchment area value, 
which describes the total number of cells that drain into that cell, 
in the next time step.
The upper bounds of the input 
$a$ is 
$b=8000$
for this research and was manually tuned. 
Thus 
$p(a)$
is the scaled percentage chance for the agent to step into the body of water. 
If the check failed, 
then the agent would walk around the obstacle by turning left or right with equal probability.
In a real-world scenario,
this equates to a hiker stepping over a small stream or encountering a lake and walking along its edges.

\paragraph*{\headtobuildings and \headtotrees}
\label{sect:vector_based_behavior}
\label{sect:head2buildings}
\label{sect:head2trees}

Using the land cover ID and digital elevation maps from 
\reftbl{tbl:list_of_gis_maps}, 
the agent could
\tech{see} the landscape and act accordingly. 
Using a viewshed algorithm means
that expensive ray-casting can be mitigated. Viewshed algorithms are popular in
GIS\cite{haverkort_computing_2009}, for applications such as radio tower
positioning. The resultant map is a mask that outlines which areas within a
given radius are visible to the observer. From this mask, any given map can be
easily analyzed using an element-wise AND operation. The visible cells can then
be analyzed further. 
To differentiate between building- and tree-seeking behavior, 
every possible land cover type is given a weight and the maximum weights are the only ones considered in the map which allows multiple map values to be treated as equals.

Then, the mean angle, $\mean{\theta}(\vec p)$, to every $c_i$ in the
set of visible cells $\vec c$ is selected as the command direction as seen below
in \refeqn{eqn:mean_angle}.

\begin{equation}	
	\mean{\theta}(\vec p) = \frac{1}{N_{vis}}\sum^{N_{vis}}_{i=0} \tan^{-1}\left(\frac{p_y - c_{i,y}}{p_x - c_{i,x}}\right)
	\label{eqn:mean_angle}
\end{equation}
where $ \vec p$ is the current position, and $N_{vis}$ is the number of visible cells. 

The weights from \reftbl{tbl:viewshed_weights} were based on the impedance values from
\citet{doherty_analysis_2014}, 
\citet{brentfrakes_national_2015}, 
and the associated behavior's intended land cover category location found. 
This is done to reflect the decision-making process that an LP might undergo whilst being lost. 
The core behavior is encoded within the weights, 
but subtle preferences between land cover IDs are also included. 
For example, 
the LP will avoid a bog at all costs whilst aiming for the arable and horticulture 
(i.e.~farmland)
if the core land cover ID target is not visible.
Given more future data, these weightings can be created for individual profiles.
\begin{table}[tb]
	\centering														
	\caption{
		The viewshed behavior weights used in this research for the building and tree behavior. 
		\headtowater was included to show how this behavior could be done without the data from 
		\citet{morris_digital_1990}.
	}
	\label{tbl:viewshed_weights}
	\begin{tabular}{@{}lrrr@{}}
		\toprule
		Land cover ID           & \headtobuildings & \headtotrees & \headtowater \\ \midrule
		Acid grassland          & 0.05             & 0.05         & 0.05         \\
		Arable and horticulture & 0.25             & 0.25         & 0.25         \\
		Bog                     & 0.01             & 0.01         & 0.01         \\
		Calcareous grassland    & 0.05             & 0.05         & 0.05         \\
		Fen, March, Swamp       & 0.03             & 0.03         & 0.03         \\
		Heather                 & 0.06             & 0.06         & 0.06         \\
		Heather grassland       & 0.05             & 0.05         & 0.05         \\
		Improved grassland      & 0.05             & 0.05         & 0.05         \\
		Neutral grassland       & 0.10             & 0.10         & 0.10         \\
		Rock                    & 0.05             & 0.05         & 0.05         \\
		Saltmarsh               & 0.05             & 0.05         & 0.05         \\
		Urban                   & 0.40             & 0.20         & 0.20         \\
		Water                   & 0.20             & 0.20         & 0.40         \\
		Woodland                & 0.20             & 0.40         & 0.20         \\ \bottomrule
	\end{tabular}	
\end{table}

\paragraph*{\headtopaths}
\label{sect:head2roads}

\newcommand{\eij}[1]{%
	\def\a{#1}\ifx\a\empty
	e_{i,j}%
	\else
	#1(\eij{})%
	\fi
}
\newcommand{\Eknearest}[0]{%
	E^{k-\text{nearest}}%
}
\newcommand{\enearest}[0]{%
	e^\text{nearest}%
}
\newcommand{\node}[1]{v_#1}

The \headtopaths behavior has two stages: \textit{hierarchical nearest path search}, and \textit{path network exploration with memory}. 
Both use a path graph where $V$ is the set of nodes, 
and $E$ is the set of edges such that 
$E \subseteq \{ \{\node{1},\node{2} \} ~| ~\node{1},\node{2} \in V~ \text{and} ~\node{1} \neq \node{2} \}$. 
Every edge in $E$ is a representation of a physical path from $\node{i}$ to $\node{j}$, 
where $\node{i},\node{j} \in V$ (shortened to $\eij{}$ for brevity) , 
that can be traversed both ways making it an undirected graph. 
Likewise, the nodes $V \in \{\node{0}, \node{1}, \dots \}$ represent the junctions or ends of paths. 
Every edge $\eij{}$ has an associated physical path $\eij{path}$ and a path type $\eij{type}$. The path type values come from the road network map in \reftbl{tbl:list_of_gis_maps} which correlates to the size of the path. This path type can then be used to map an edge to an integer score 
\begin{equation}
	\eij{score} = 
	\begin{cases}
		10, & \eij{type} = \text{Major road} \\
		5,  & \eij{type} = \text{Trunk road} \\
		2,  & \eij{type} = \text{Path}       \\
		4,  & \text{else}                             
	\end{cases}
	\label{eqn:method_to_convert_road_hiera_id_to_score}
\end{equation}

\begin{figure*}[htb!]
	\centering
	\subfloat[Nearest points on 3 paths from $(0,0)$ (red) where each path has an associated score representing $score(\Eknearest)$ where $k=3$.\label{fig:nearest_points_paths}]{\includegraphics[width=0.32\linewidth]{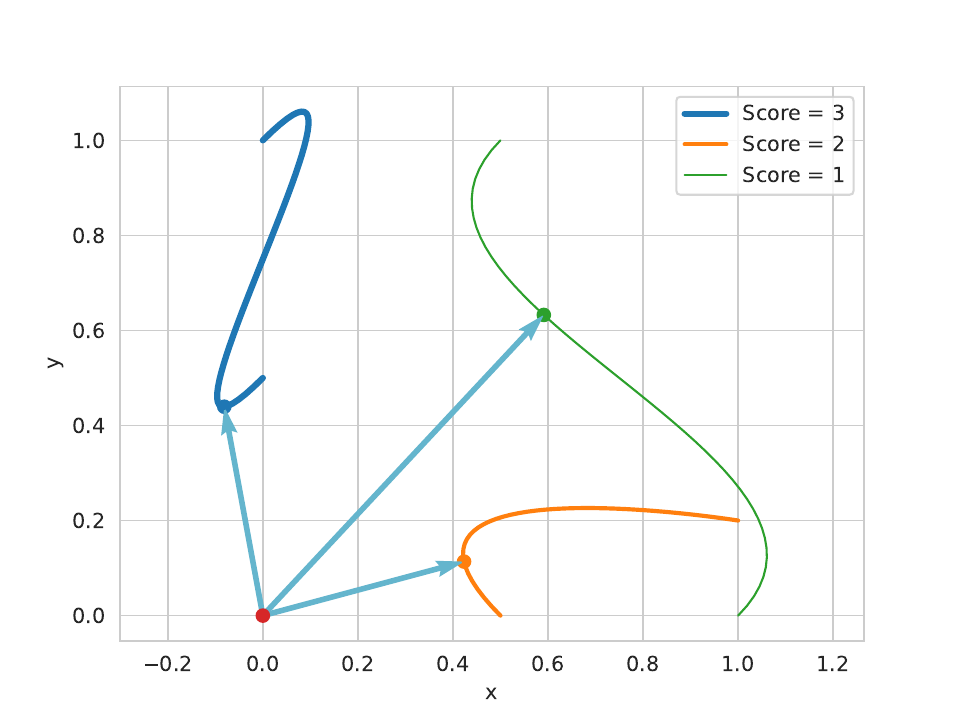}}
	\hfill
	\subfloat[Scaled vectors based on \refeqn{eqn:dist_over_score_scale} and \reffig{fig:nearest_points_paths}.\label{fig:weighted_nearest_points_vec}]{\includegraphics[width=0.32\linewidth]{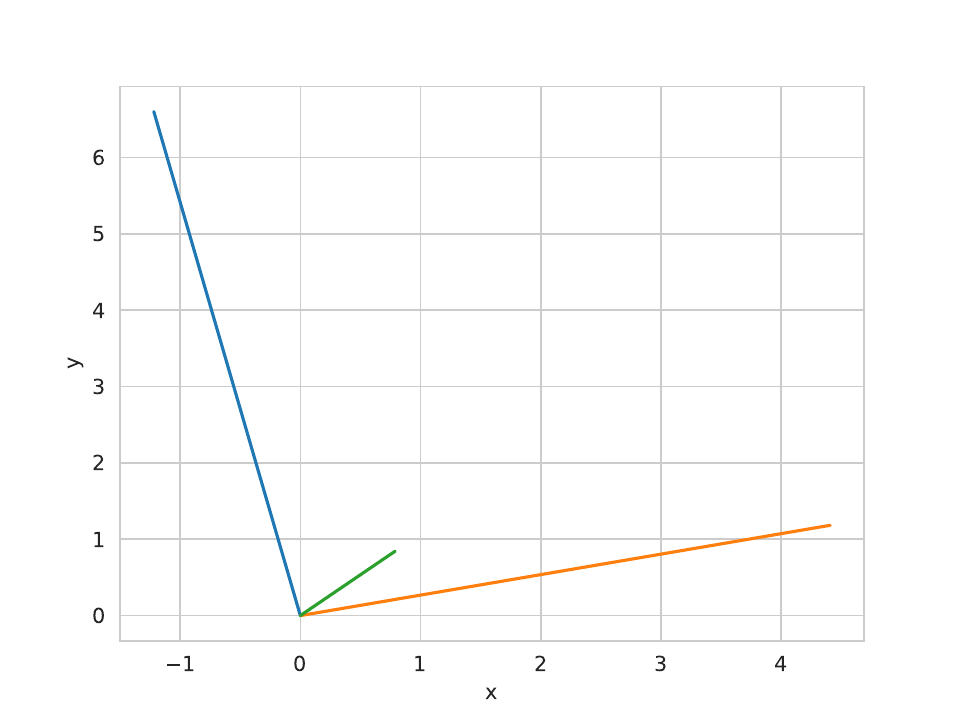}}
	\hfill
	\subfloat[Tip-to-tail stacked scaled vectors from \reffig{fig:weighted_nearest_points_vec} resulting in weighted average vector from which the angle is calculated.\label{fig:weighted_average_angle}]{\includegraphics[width=0.32\linewidth]{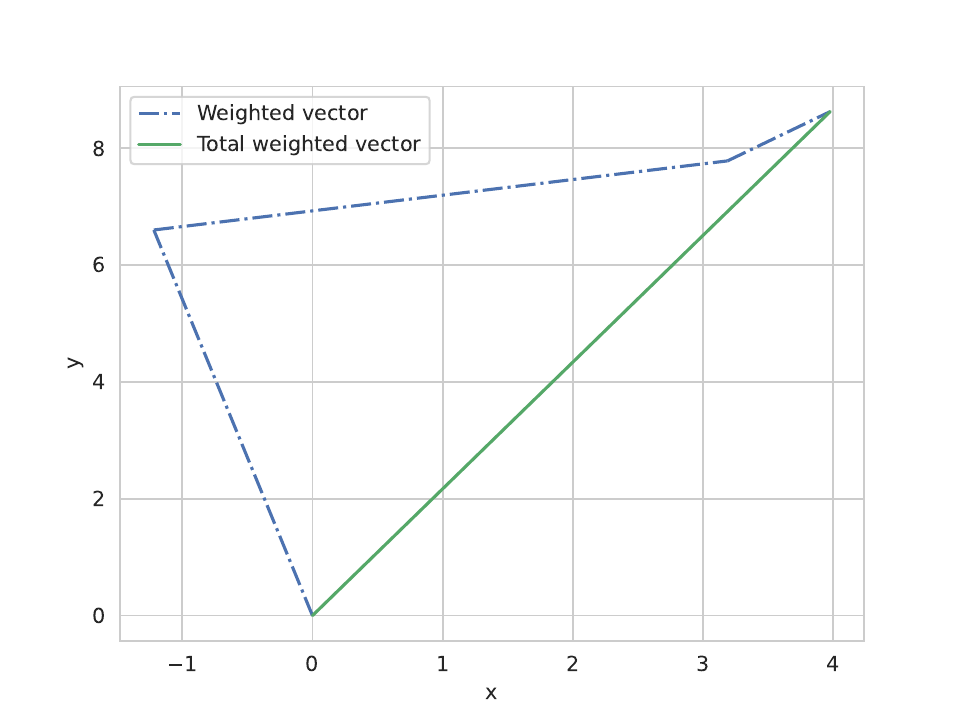}}
	\caption{The process of extracting the weighted average angle $\mean\theta$ from $\Eknearest$ for a given position $\vec m $ on the grid.}
	\label{fig:process_to_get_weighted_average_angle}
\end{figure*}

Using the above-defined path graph, 
the hierarchical nearest path search then uses the geospatial information of the path; $\eij{path}$
and 
$\eij{score}$.
On an evenly spaced grid 
$\vec M$
of points covering the search area,
every
$\vec m \in \mathbb{R}^2$
has 
$k$
distinct nearest edges
$\Eknearest 
\subset E
$.
This subset is found by finding the Euclidean distance from 
$\vec m$
to the single closest location on every edge's path
$\eij{point} \in \mathbb{R}^2$
and then selecting the $k$ closest ones.
The weighted average angle 
$\mean\theta$
that the agent will move in is then
\begin{gather}
	d_n = ||\vec m - point(\Eknearest_n)|| \\
	c_n = \frac
		{score(\Eknearest_n)}
		{d_n}
		\label{eqn:dist_over_score_scale}\\
	\hat u_n =  \frac{point(\Eknearest_n)-\vec m}{d_n} \\
	\mean\theta = \arctan\left(\sum_{n=0}^{k} c_n \hat u_n\right)\label{eqn:multi-path-mean-angle}
\end{gather}
where $c_n$
is the scaling term associated to
$\Eknearest_n$,
and 
$\hat u_n$ 
is the unit vector in the direction from 
$\vec m$
to
$point(\Eknearest_n)$.
This process can be seen in 
\reffig{fig:process_to_get_weighted_average_angle},
showing the influence that the scaling introduced by
\refeqn{eqn:method_to_convert_road_hiera_id_to_score} 
and 
\refeqn{eqn:dist_over_score_scale} 
has on the weighted average angle.
As the agent traverses the landscape looking for a path, 
they follow the angle associated with their current position. 
The weighted nature of the paths in 
\refeqn{eqn:multi-path-mean-angle} 
represents the agent being able to prioritize a major road over a small dirt path, 
but still giving precedence over a smaller, 
closer, 
walking path via the 
$\frac{1}{d}$ 
scaling in 
\refeqn{eqn:dist_over_score_scale}.

\begin{figure}[htb!]
	\centering
	\includegraphics[width=0.9\linewidth]{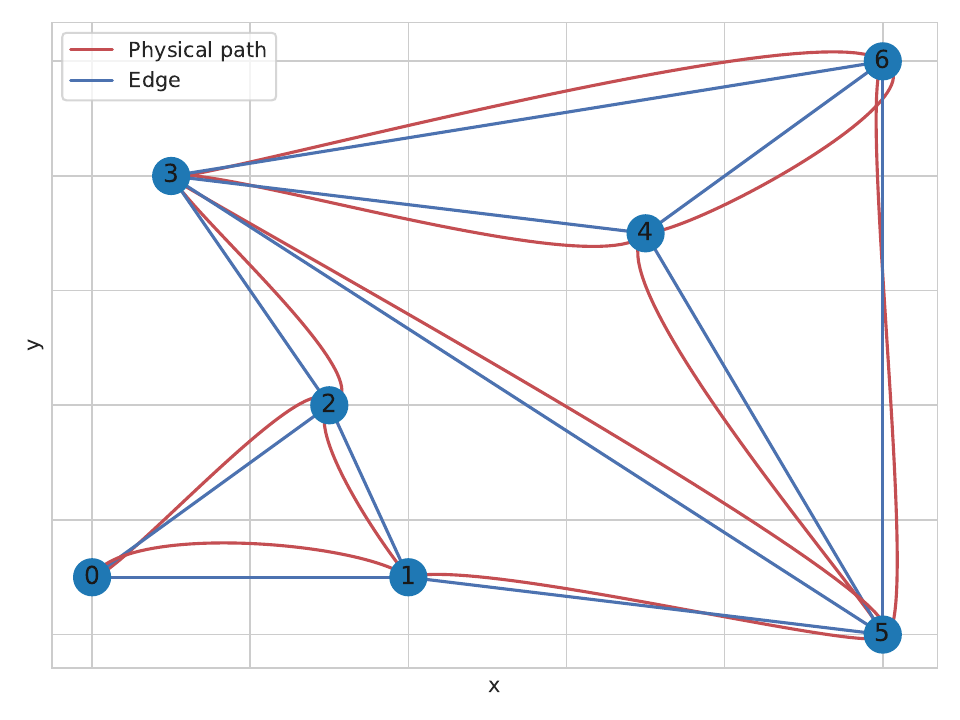}
	\caption{Example path network. If the agent moves from $0$ to $2$, the edge $(0,2)$ will have a score of $0.1$ assigned to it whilst both $(2,3)$ and $(2,1)$ will have a score of $1$. After normalizing the scores, this gives a probability of $4.76\%$ to backtrack to $0$.}
	\label{fig:example_path_network}
\end{figure}

Once the agent reaches a path, the behavior switches to traversing the path network, as seen in \reffig{fig:example_path_network}.
At this switching position, $\vec x$, the nearest path is searched for, as outlined above but with $k=1$.
Then, the point $\vec x$ is inserted into $V$ as a new node $\node{\text{new}}$ and two new edges, 
$\enearest_- = \{ \node{\text{new}}, \node{i} \}$ and $\enearest_+= \{ \node{\text{new}}, \node{j} \}$, 
are inserted into $E$ and the associated physical paths undergo a similar process.
The original path $path(\enearest)$ is cut at $\vec x$ into two pieces.
One piece is reversed to start at $\vec x$, and the other is left as is, but a straight line from $\vec x$ to the path is inserted for completeness.
These new edges are marked as being non-traversable once the agent has moved along them.
The agent then begins traversing the path network from node $\node{\text{new}}$ by
selecting randomly from $\enearest_-$ and $\enearest_+$ with equal probability. This physically
represents a person coming across a path and either turning left or right.

At every subsequent step, the agent moves along the whole physical path $\eij{path}$ entirely before the next maximum distance check is done. As the upper time limit is observed during the sampling stage, this does not impact the final result and improves performance by not requiring costly spatial computations on the paths. 
Once it reaches the next node, the agent randomly selects the next edge from the node's adjacency list, 
which is a finite graph representing the set of neighbors of that particular node.
Every edge in the adjacency list is assigned a score of being selected.
It is given a score of 0.1 if it has been visited in the last 
$\lambda_{\max}$ 
steps, 
or 1 if it hasn't.
This vector of scores is normalized to give a probability vector which is then used to randomly select an edge to traverse.
This discourages the agent from backtracking, 
but does not completely forbid it which is crucial to prevent the agent from becoming isolated on a sub-graph of the network.

\subsection{Sampling Of Paths}
\label{sect:method_sampling}

\begin{figure}[htb!]
	\centering
	\includegraphics[width=0.9\linewidth]{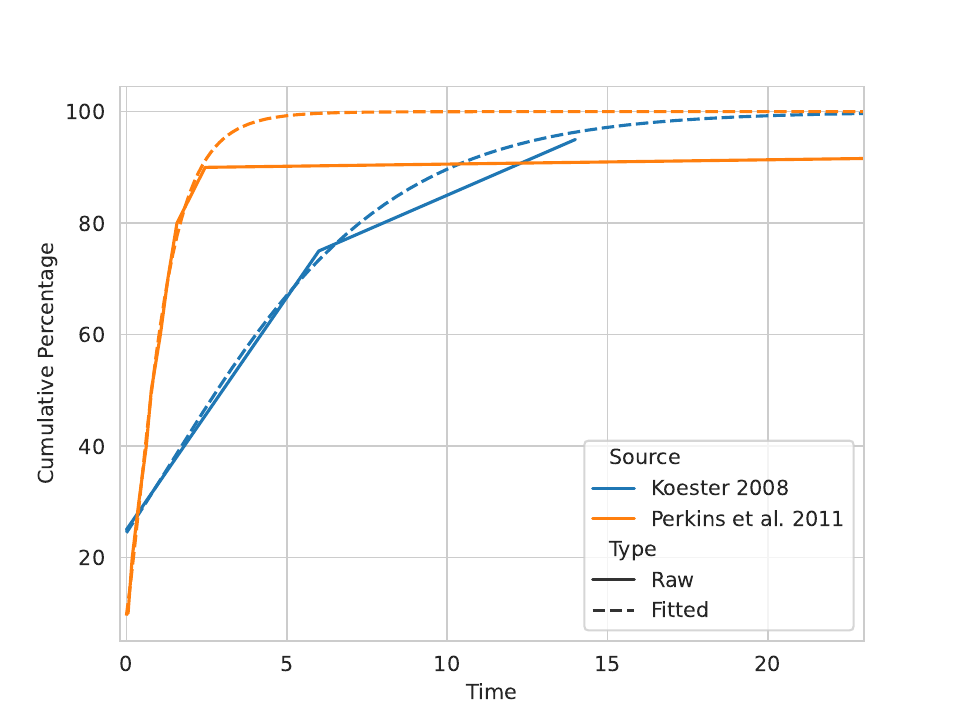}
	\caption{Discrete mobility time (hours) from both \citet{koester_lost_2008} (n=232) and \citet{perkins_missing_2011} (n=132) along with their respective continuous Log-Normal curve. \citet{perkins_missing_2011} provides distance from PLS data in kilometers, and was scaled by $3.87\si{\kilo\metre\per\hour}$ from \citet{gast_preferred_2019}}
	\label{fig:koester_and_perkins_mobility_time}
\end{figure}

The simulation length has a termination distance of $10,000\si{\m}$, which gives a simulated time of $2.58\si\hour$, and accounts for over $95\%$ of LP scenarios when using the \citet{perkins_missing_2011} dataset. \reffig{fig:koester_and_perkins_mobility_time} shows the two mobility time distributions found from historical data which measures the amount of time the LP was moving\cite{koester_lost_2008,perkins_missing_2011}.

To fit the mobility time distribution, multiple other functions were evaluated: exponential, log-gamma, and normal. However, by using the symmetrical Kullback-Leibler divergence function\cite{kullback_information_1951}

\begin{equation}
	SKL(P||Q) = \sum p_i(x)\log\left(\frac{p_i(x)}{q_i(x)}\right) + \sum q_i(x)\log\left(\frac{q_i(x)}{p_i(x)}\right)\label{eqn:symmetric-kl}                            
\end{equation}
	
as a metric, it was found that the log-normal function\cite{johnson_continuous_1994} was the best fitting. Where the log-normal is defined as
	
\begin{equation}
	\begin{gathered}
		f(y) = \frac{1}{s\cdot y\cdot\lambda\sqrt{2\pi}}\exp{\left(  -\frac{\log^2(y)}{2s^2}  \right)}
		\\
		y \mapsto \frac{x-\mu}{\lambda}
	\end{gathered}
	\label{eqn:lognorm}
\end{equation}
	
where $s$ is the shape parameter, $\mu$ is the mean, and $\lambda$ is the scaling parameter.

To sample the location found from the $N_{gen}$
generated paths from section \ref{sect:lp_behavior},
$M$ time samples were taken from 
\refeqn{eqn:lognorm} for every path.
A constant speed of
$3.87\si{\hour\per\kilo\metre}$,
which is the average preferred walking speed of a hiker over rough terrain\cite{gast_preferred_2019}, 
is used to convert time to distance.
A given path is then sampled at the given distances along it to generate the locations found.
This is repeated for all $N_{gen}$ paths to generate $N_{gen} \cdot M$ samples.

\section{Design Of Experiment}
\label{sect:doe}

\begin{figure}[htb!]
	\centering
	\includegraphics[width=0.9\linewidth]{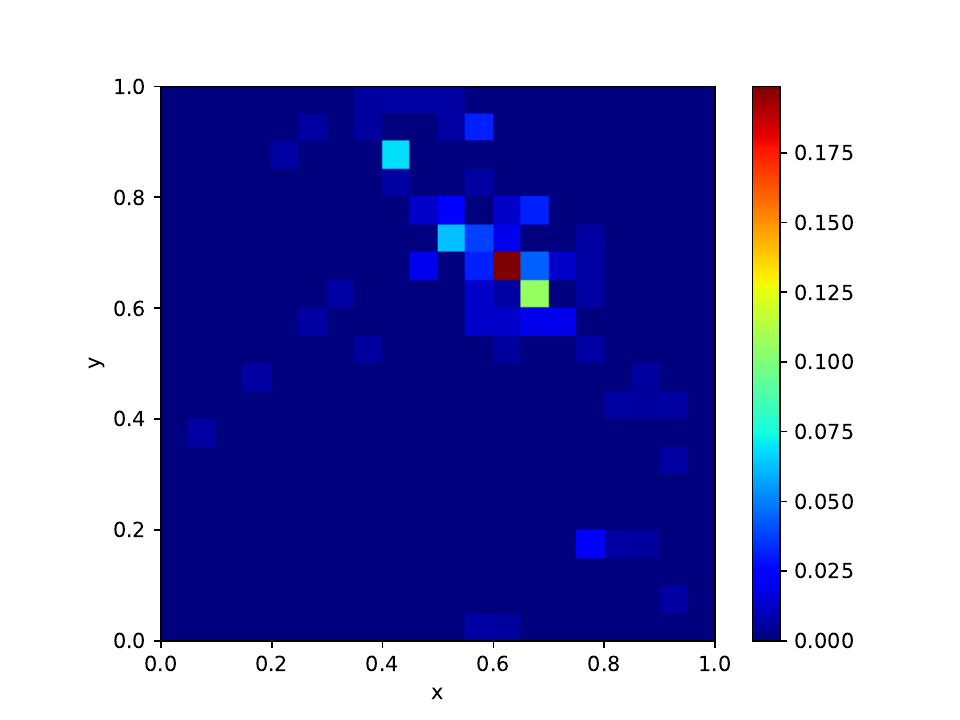}
	\caption{Normalized ($x$-,$y$- and $z$-axis) heatmap derived from the raw SMR PLS data}
	\label{fig:pls_raw_data_heatmap}
\end{figure}

The data supplied by \citet{perkins_missing_2011}, \citet{koester_lost_2008}, and
\citet{grampianpolice_missing_2007} combines generalized data from hundreds of LP
cases to draw broad conclusions about the various LP profiles.
Therefore, to draw similar conclusions from the methods outlined in
\refsect{sect:method}, an adequate distinct number of PLS locations must be used.
SMR provided a sample dataset for the Isle of Arran of
historical rescue data, with $\approx300$ PLS locations. This position data is
in the form of a 6-digit Ordnance Survey National Grid reference, which reduces
the British Isles into $100 \times 100 \si{\m}$ cells. A PLS described by this system has a $\pm 100 \si\m$ error in $x$ and $y$, with a total magnitude of the error being $\pm 141.42 \si{\m}$. This means a
heatmap generated from this data has a cell size of $100\si\m$, as seen in
\reffig{fig:pls_raw_data_heatmap}. This would result in a large cluster of
sampled points around $(0.6,0.7)$, canceling out the attempts at increasing the
number of distinct locations.

\begin{figure}[htb!]
	\subfloat[Bicubic\cite{getreuer_linear_2011}\label{fig:pls_bicubic_sampled_heatmap}]{\includegraphics[width=0.9\linewidth]{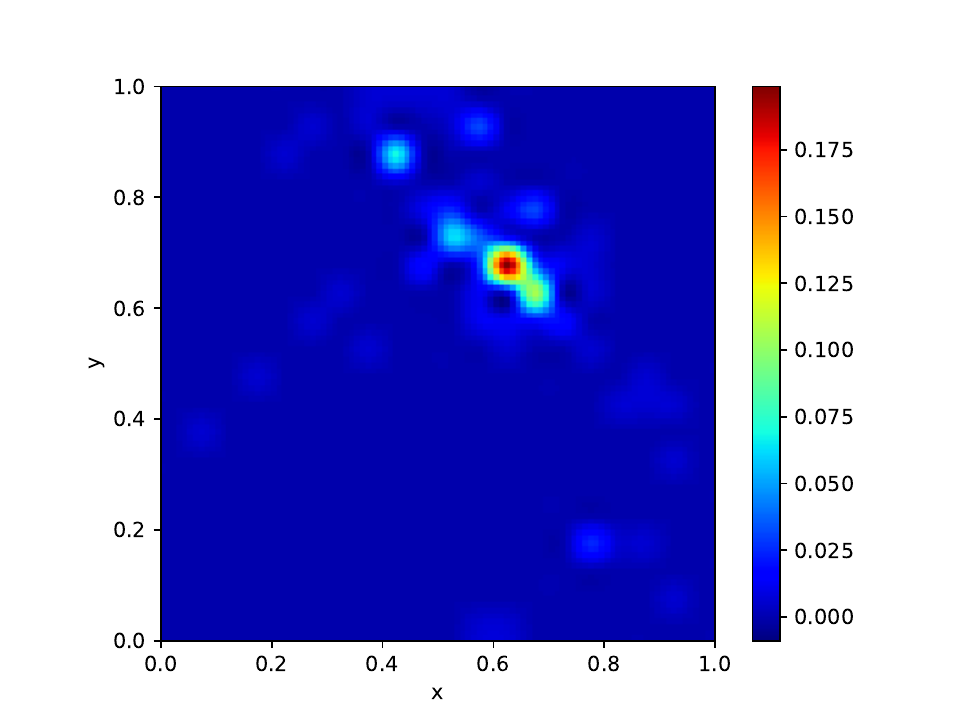}}
	\hfill
	\subfloat[Lanczos\cite{getreuer_linear_2011}\label{fig:pls_lanczos_sampled_heatmap}]{\includegraphics[width=0.9\linewidth]{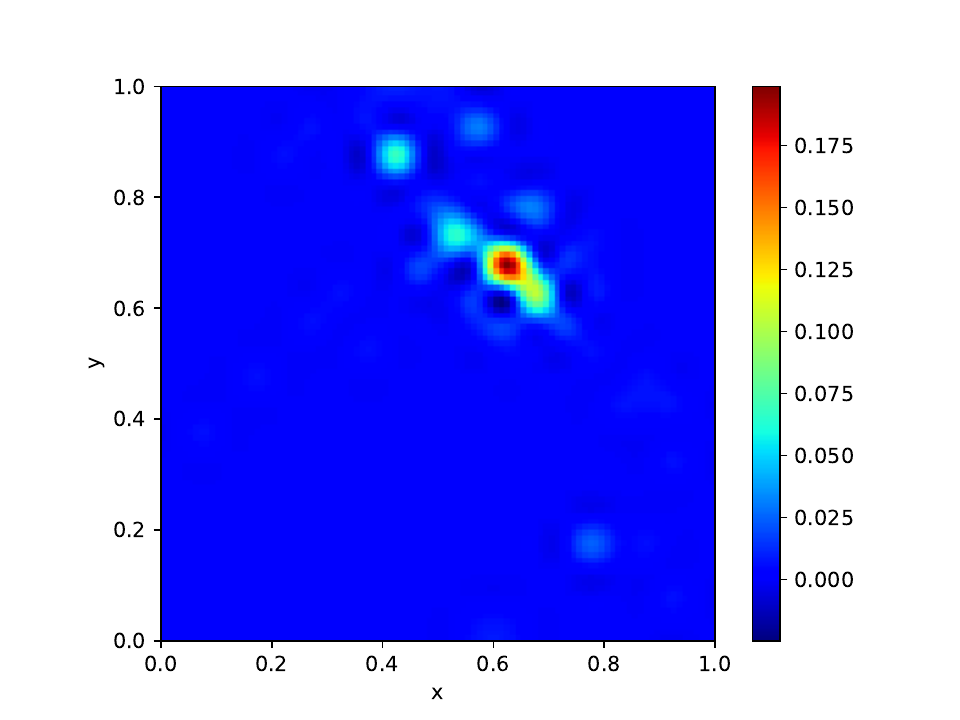}}
	\caption{The resultant heatmaps after applying conventional methods to \reffig{fig:pls_raw_data_heatmap}}
	\label{fig:pls_other_upsampling_methods}
\end{figure}
To up-sample the resolution of the heatmap seen in \reffig{fig:pls_raw_data_heatmap},
fitting a continuous function to represent the probability is a suitable solution. 
Using other methods, 
such as bicubic or Lanczos interpolation\cite{getreuer_linear_2011},
results in the same disconnected modes as the original heatmap. 
This can be observed in 
\reffig{fig:pls_other_upsampling_methods}. 
To address the issues with these methods we used supervised machine learning in the form of a
\acrdef{Gaussian Process}{GP} model. 
The results of this method are discussed in 
\refsect{sect:doe_results}.
This approach is appropriate for noisy and uncertain systems\cite{rasmussen_gaussian_2006}. 
A GP learns a prior over functions,
which can be sampled after observing data.

For the GP to work efficiently, the data must be formatted correctly.
The original heatmap 
$z^-$
is normalized through
\begin{equation}
	z^+
	=
	\frac
	{z^--z^-_\text{min}}
	{z^-_\text{max}-z^-_\text{min}}
	\label{eqn:gp_data_normalization}
\end{equation}
to give 
$z^+ \in [0,1]$. 
The $x$ and $y$ values are normalized in the same manner. 
The normalized 2D heatmap is then unraveled in the row-major form such that every unique
$(x, y)$ 
pair maps to a single 
$z^+$
value in a 1D array.

The GP is then based primarily on the Matern kernel
\cite{rasmussen_gaussian_2006} which computes a covariance matrix between inputs
$\mathbf{a}$ and $\mathbf{b}$:

\begin{equation}
	\begin{gathered}
		k_{\text{Matern}}(\mathbf{a}, \mathbf{b}) 
		= 
		\frac
		{2^{1 - \nu}}
		{\Gamma(\nu)}
		\left( \sqrt{2 \nu} d \right)^{\nu} 
		K_\nu \left( \sqrt{2 \nu} d \right) \\
		d = (\mathbf{a} - \mathbf{b})^\top \Theta^{-2} (\mathbf{a} - \mathbf{b})
	\end{gathered}
\end{equation}

where $d$ is the distance between $\mathbf{a}$ and $\mathbf{b}$, 
scaled by the length scale parameter $\Theta$, 
$\nu$ is a smoothness parameter\cite{gardner_gpytorch_2021}, 
and $K_\nu$ is the modified Bessel function\cite{abramowitz_handbook_1972}.
$k_\nu$ can be simplified at half-integer steps of $\nu$ 
($\nu = p + \frac{1}{2} ~ \forall ~ p \in \mathbb N$) 
with the most commonly used values being $\nu = 1.5$ and $\nu = 2.5$. 
This is due to $\nu=0.5$ generally giving noisy outputs, and $\nu \geq 3.5$ being hard to distinguish between\cite{rasmussen_gaussian_2006}.
As such,
a value of $\nu=2.5$\cite{cornford_modelling_2002} was select which simplifies the modified Bessel function to

\begin{equation}
	k_{\nu=2.5}
	=
	\exp
	\left(1+\frac{\sqrt{3}r}{l}+\frac{5r^2}{3l^2}\right)
	\left(\exp -\frac{\sqrt{5}r}{l} \right)
\end{equation}

The Matern kernel is then scaled using a scale kernel such that 

\begin{equation}
	K_{\text{scaled}} = \theta_\text{scale} K_{\text{orig}}	
\end{equation}

where, $\theta_\text{scale}$ is the output scale. As the dataset is image-based,
and has a large amount of data points as a result, the KISS-GP
\cite{wilson_kernel_2015} approximation for a given kernel is then applied.
Given a base kernel $K$, the covariance $k(\mathbf{a}, \mathbf{b})$ is
approximated by using a grid of regularly spaced \textit{inducing points}:

\begin{equation}
	k(\mathbf{a}, \mathbf{b}) = \mathbf{w_{a}}^\top K_{U,U} \mathbf{w_{b}}
\end{equation}

where $U$ is the set of gridded inducing points, $K_{U,U}$ is the kernel matrix
between the inducing points, $\mathbf{w_{a}}$ and $\mathbf{w_{b}}$ are sparse
vectors based on $\mathbf{a}$ and $\mathbf{b}$ that apply cubic interpolation.

This covariance is then combined with a zero mean, in a multivariate normal
distribution, and quantified by an exact marginal log-likelihood. This allows
the measurement of the probability of generating the observed sample from a
prior\cite{gardner_gpytorch_2021}. Finally, this is optimized using the Adam
optimization algorithm\cite{kingma_adam_2017}.

\section{Results and Discussion}
\label{sect:results_and_discussion}

\subsection{Experiment PLS Locations}
\label{sect:doe_results}

A key element in deciding which location to analyze was the availability of the historical PLS data outlined in \refsect{sect:doe}. 
The availability of that data, coupled with the availability of a high-quality LiDAR digital elevation map\cite{ascgeospatialdata_lidar_2014}, 
meant that the Isle of Arran,
Scotland, 
was therefore chosen. 

\begin{figure}[htb!]
	\centering
	\includegraphics[width=0.9\linewidth]{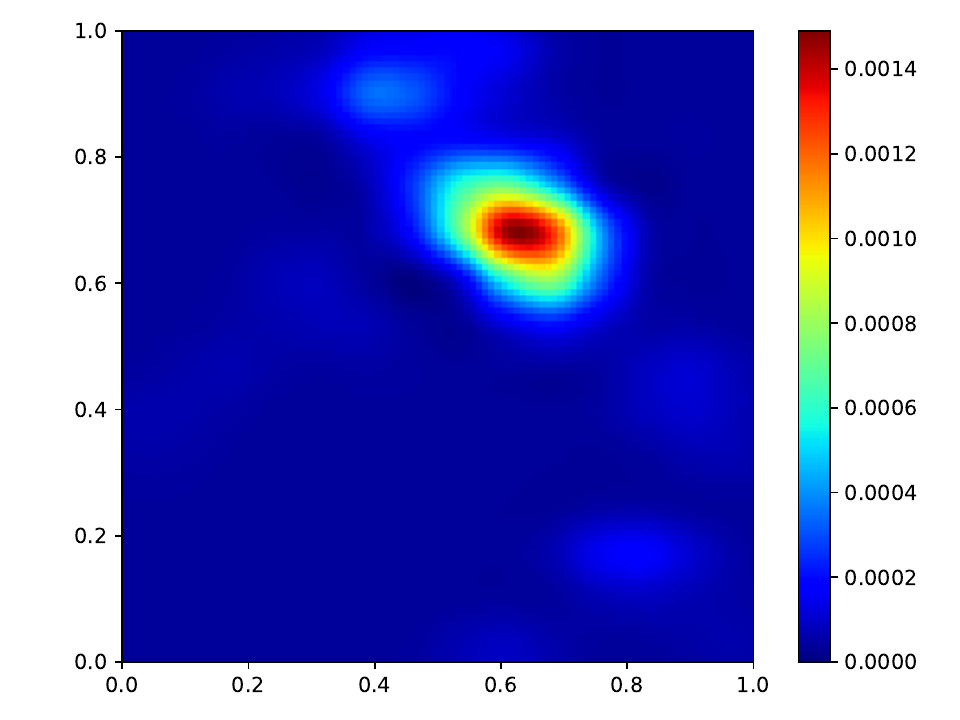}
	\caption{Up-sampled \reffig{fig:pls_raw_data_heatmap} using a Gaussian
		Process. Every grid cell now represents a $20\times20\si\m$ area resulting
	in more possible sample locations.}
	\label{fig:pls_gp_sampled_heatmap}
\end{figure}

The purpose of the process outlined in \refsect{sect:doe} was to estimate the raw data provided by SMR,
and sample PLS locations from this model to further drive the experiments such that
generalized statistics could be drawn from it.
\reffig{fig:pls_gp_sampled_heatmap} shows the up-sampled heatmap, which retains
the geospatial information of the PLS locations whilst increasing the number of
sample points infinitely through the GP model.

\begin{figure}[htb!]
	\centering
	\includegraphics[width=0.9\linewidth]{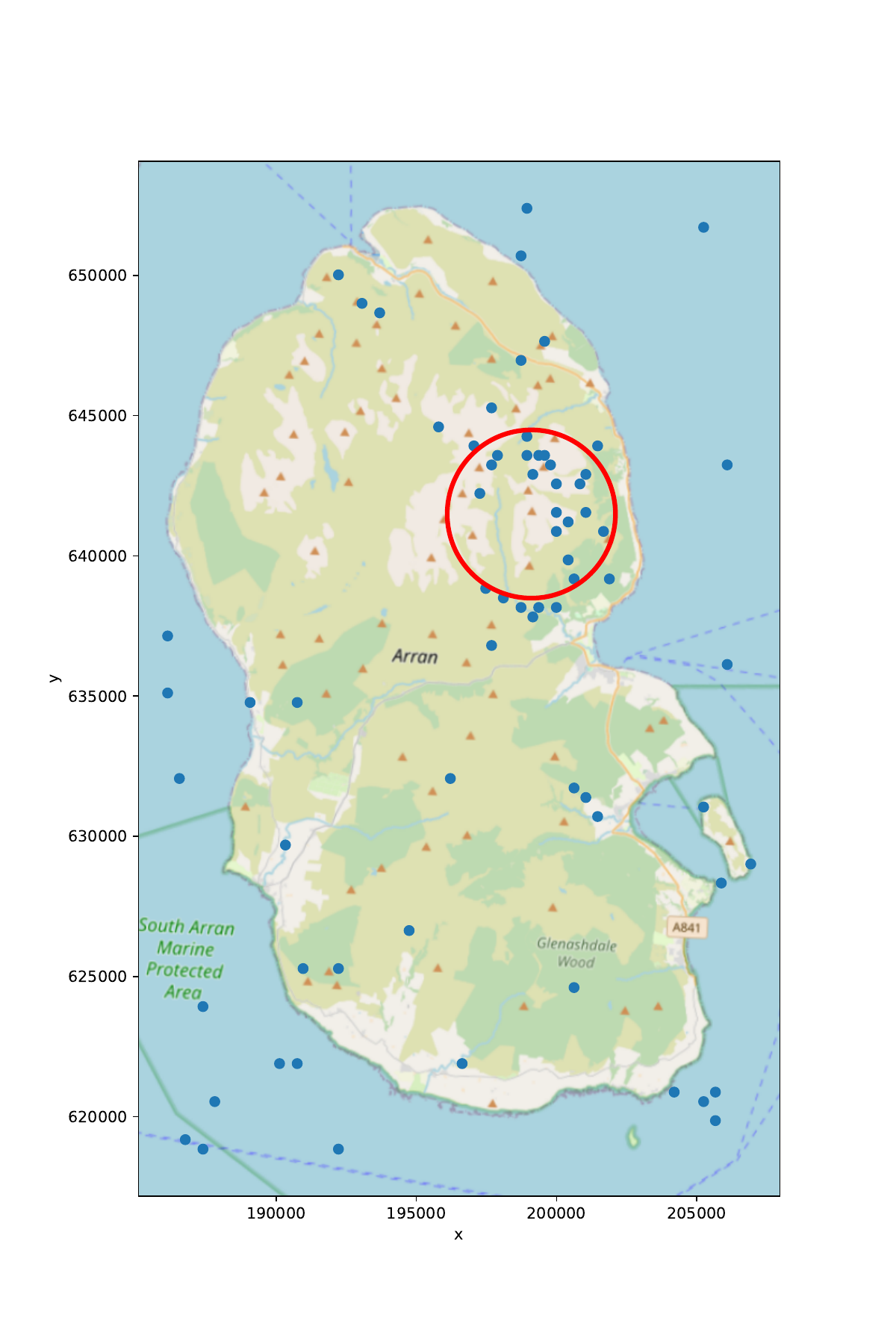}
	\caption{$n=70$ sampled locations from \reffig{fig:pls_gp_sampled_heatmap}
		with the normalization of \refeqn{eqn:gp_data_normalization} reversed.
		Note the cluster of PLS locations around the hotspots (circled in red) from the raw data seen in \reffig{fig:pls_raw_data_heatmap} with a base map from \citet{openstreetmapcontributors_openstreetmap_2017}} 
	\label{fig:pls_70_sampled_points_over_arran}
\end{figure}

\reffig{fig:pls_70_sampled_points_over_arran} shows a small sample of $n=70$
points ($n$ was kept low for better figures) taken
directly from the GP model. The nature of the GP model ignores the geospatial
constraints of hikers not getting lost in the middle of the sea. Dealing with
this is a task for the PDM generation algorithm.

The start locations were then sampled from the trained GP model, like in \reffig{fig:pls_70_sampled_points_over_arran}, and then further sampled from \refeqn{eqn:gaussian} with 
$\sigma_{xx} = \sigma_{yy} = 10,000$
to model the uncertainty in the reported PLS.
In total 
$270$
starting points were evaluated resulting in 
$50,000$ 
generated paths. 
The final number of locations found was 
$41,000,000$
after the sampling stage.
This is a sufficiently large number of data points for statistical analysis since
\citet{koester_lost_2008}
and
\citet{perkins_missing_2011}
reported
$n=3,800$ 
and
$n=130$ 
respectively.

\subsection{Paths}

\begin{figure*}[htb!]
	\centering
	\subfloat[\headtowater\label{fig:routes_head2water}]{\includegraphics[width=0.24\linewidth]{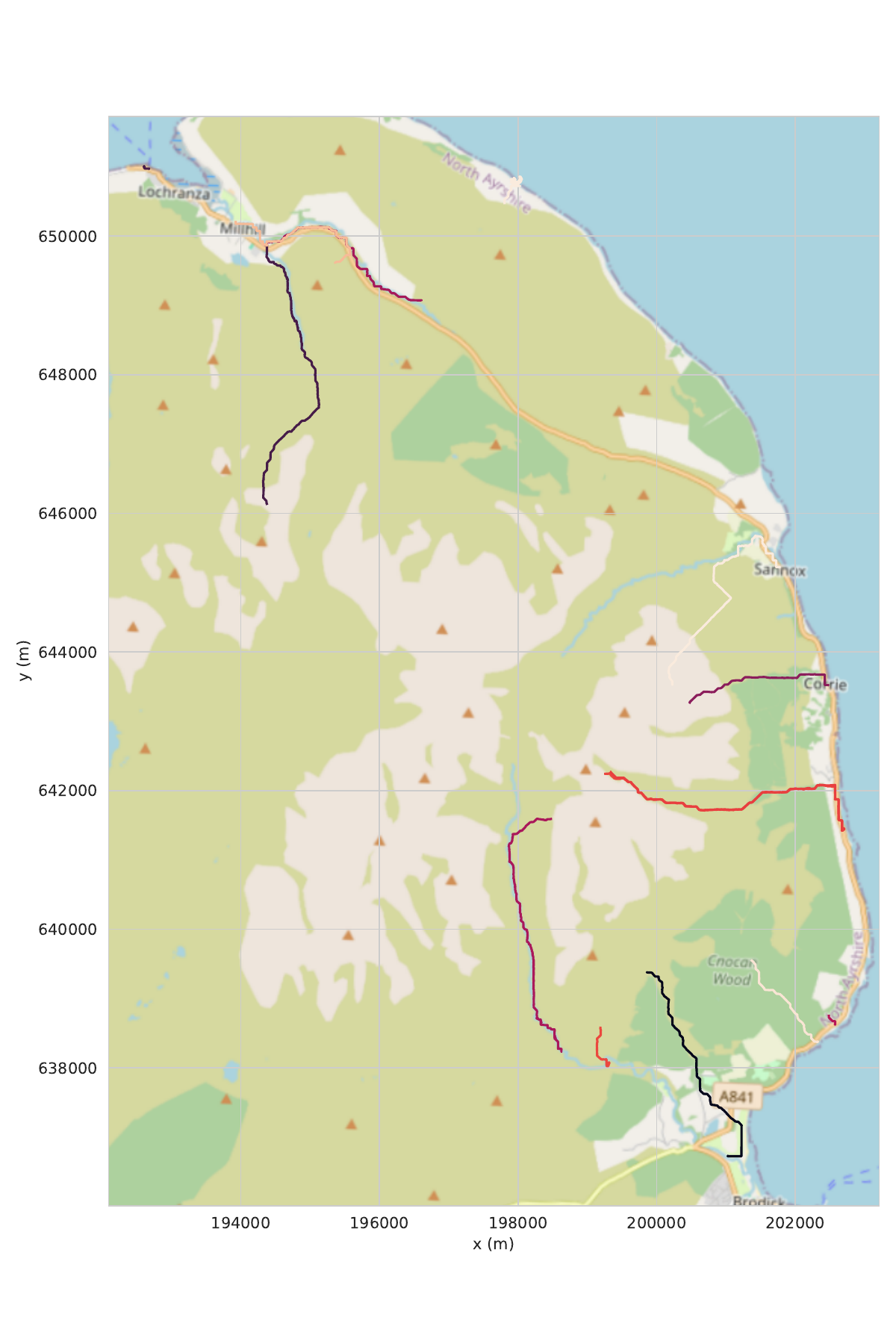}}
	\hfill
	\subfloat[\headtobuildings\label{fig:routes_head2buildings}]{\includegraphics[width=0.24\linewidth]{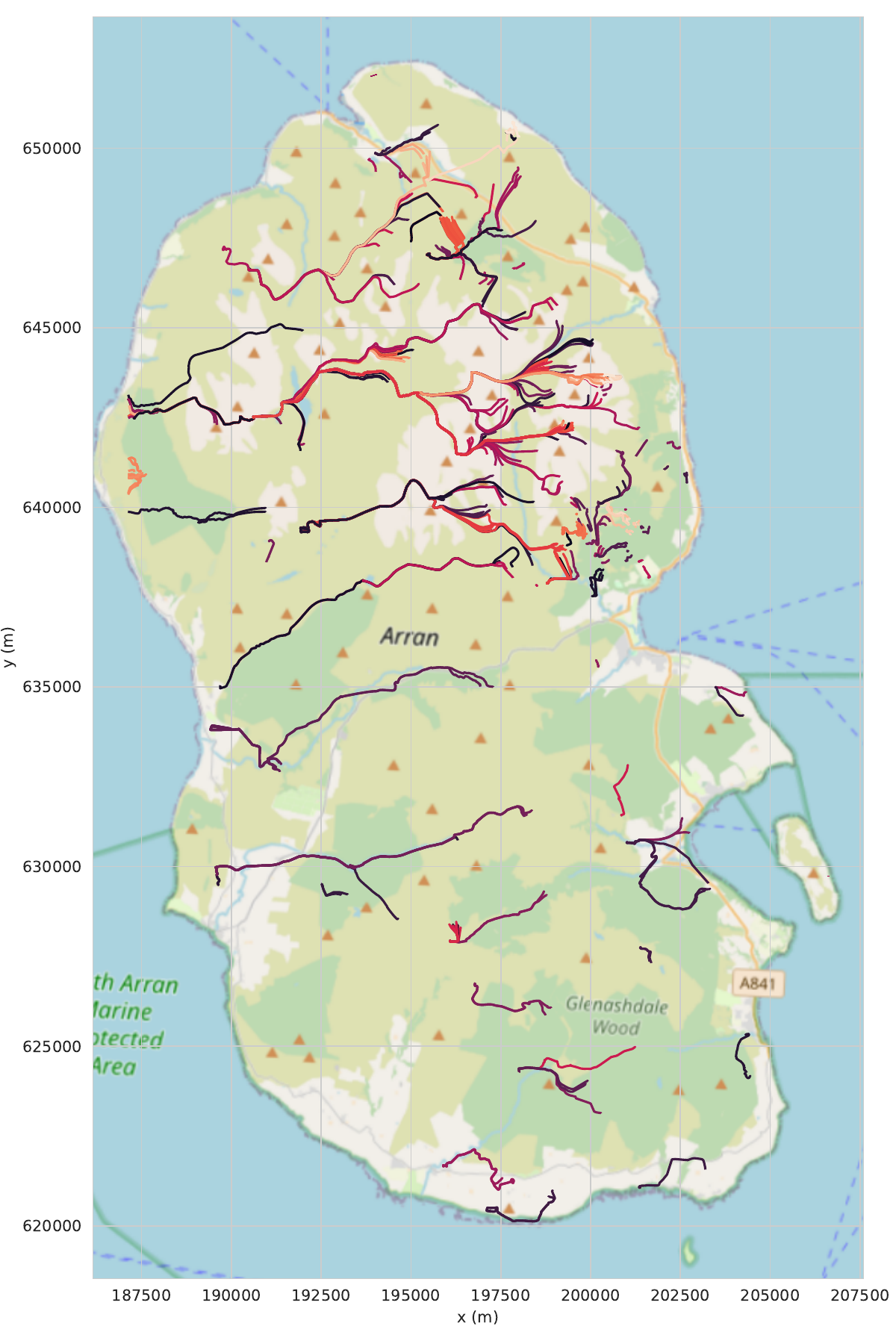}}
	\hfill
	\subfloat[\headtopaths\label{fig:routes_head2roads}]{\includegraphics[width=0.24\linewidth]{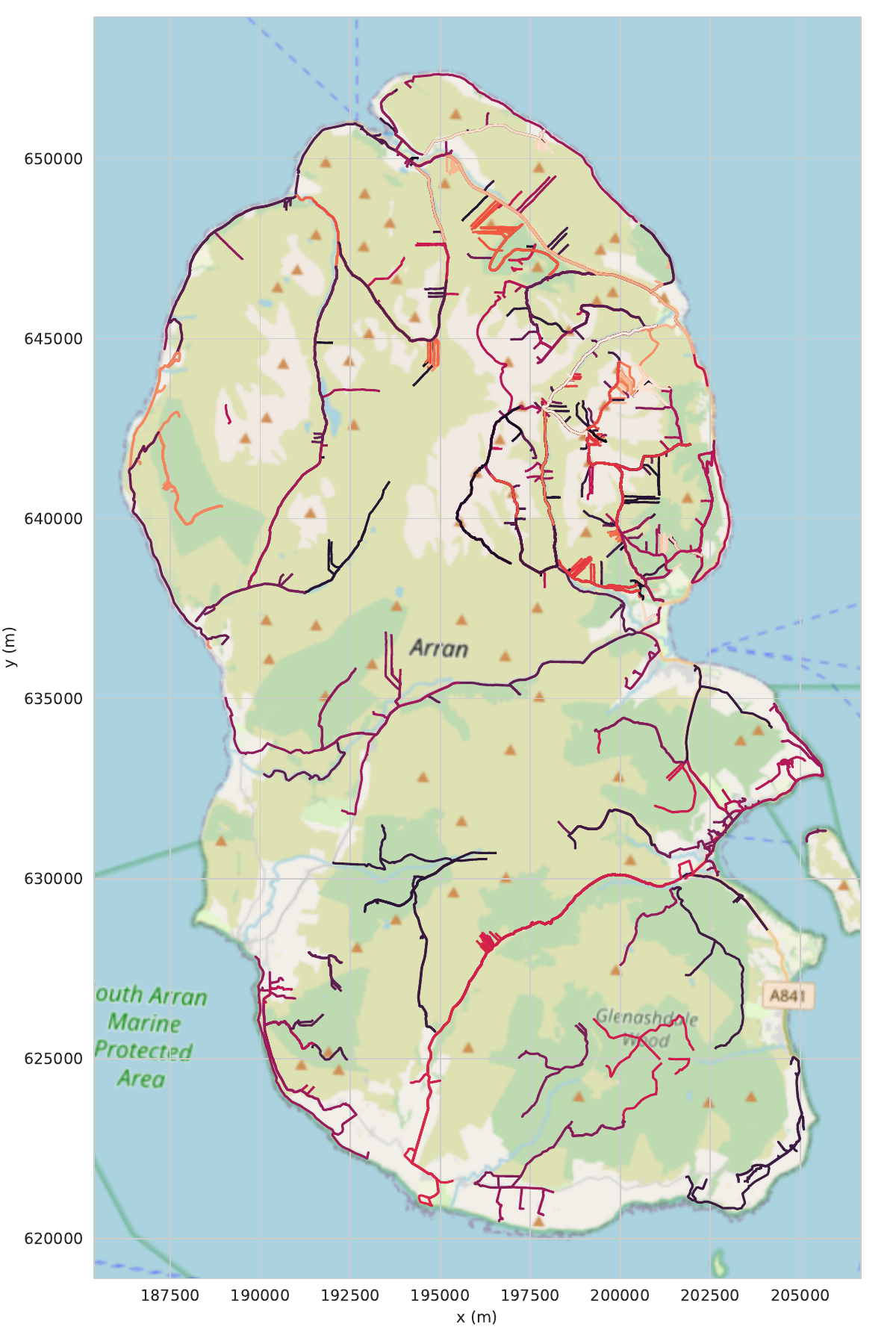}}
	\hfill
	\subfloat[\headtotrees\label{figroutes_head2trees:}]{\includegraphics[width=0.24\linewidth]{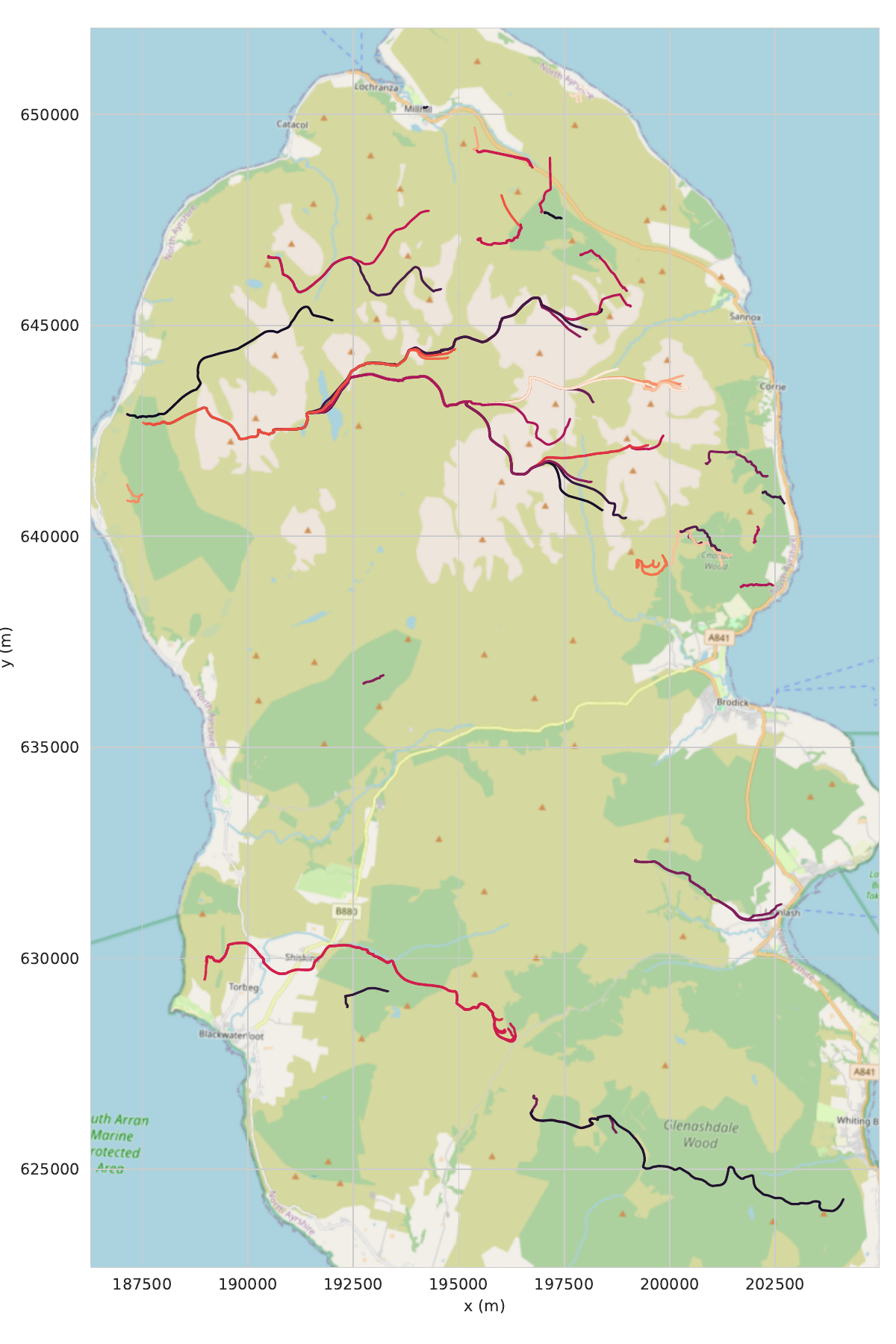}}
	\caption{Subset ($n=1,000$) of generated paths traversed by agents using the four different behaviors.
	The color is used for a contrast between paths with the base map from \citet{openstreetmapcontributors_openstreetmap_2017}.
	}
	\label{fig:behavior_routes}
\end{figure*}

\reffig{fig:behavior_routes} shows the paths created by various behaviors. As
expected, there are substantially fewer paths for \headtowater than there are
for \headtopaths. From \reffig{fig:routes_head2buildings}, it can be seen that
due to the lack of buildings in the north of the Isle of Arran, the paths there
converge on a couple of points whereas they tend to converge on different
locations in the south. Furthermore, this behavior shows the effect of the
sampled points from \refsect{sect:doe}.

\begin{figure}[htb!]
	\centering
	\subfloat[Distribution of simulated path lengths showing tendencies for some behaviors (\headtopaths, \headtowater) to travel further than others (\headtotrees, \headtobuildings)\label{fig:path_length_distribution}]{\includegraphics[width=0.9\linewidth]{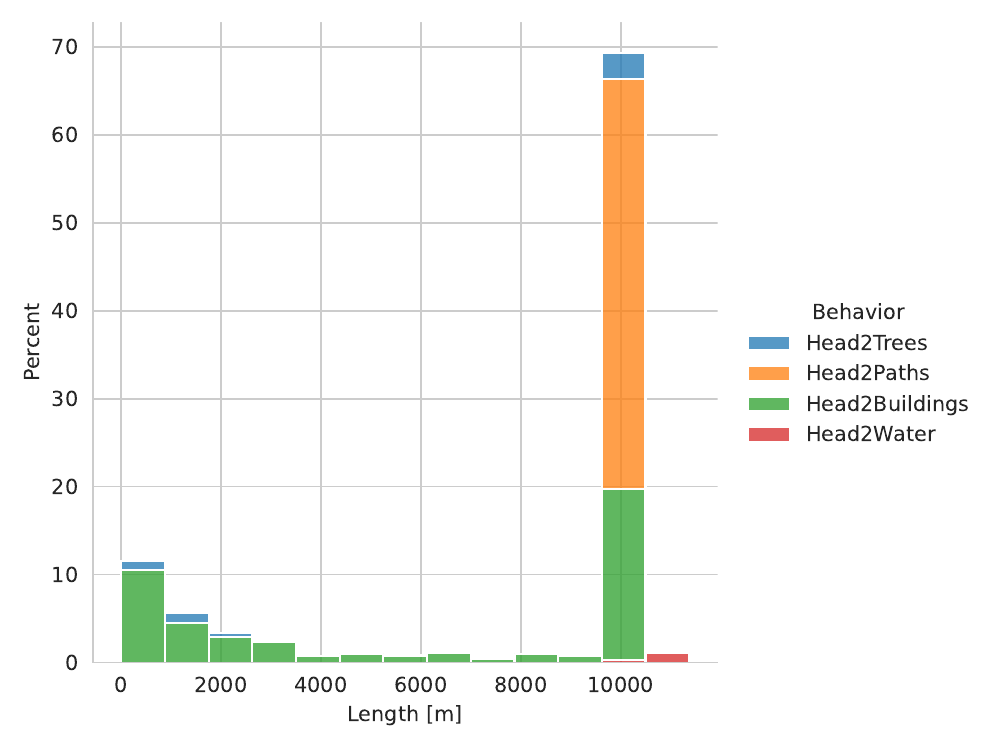}}
	\hfill
	\subfloat[Both viewshed-based algorithms are seen to have a low median around, $4.50\si{\kilo\metre}$, whilst the vector map algorithms are closer to $10.50\si{\kilo\metre}$\label{fig:path_length_boxplot}]{\includegraphics[width=0.9\linewidth]{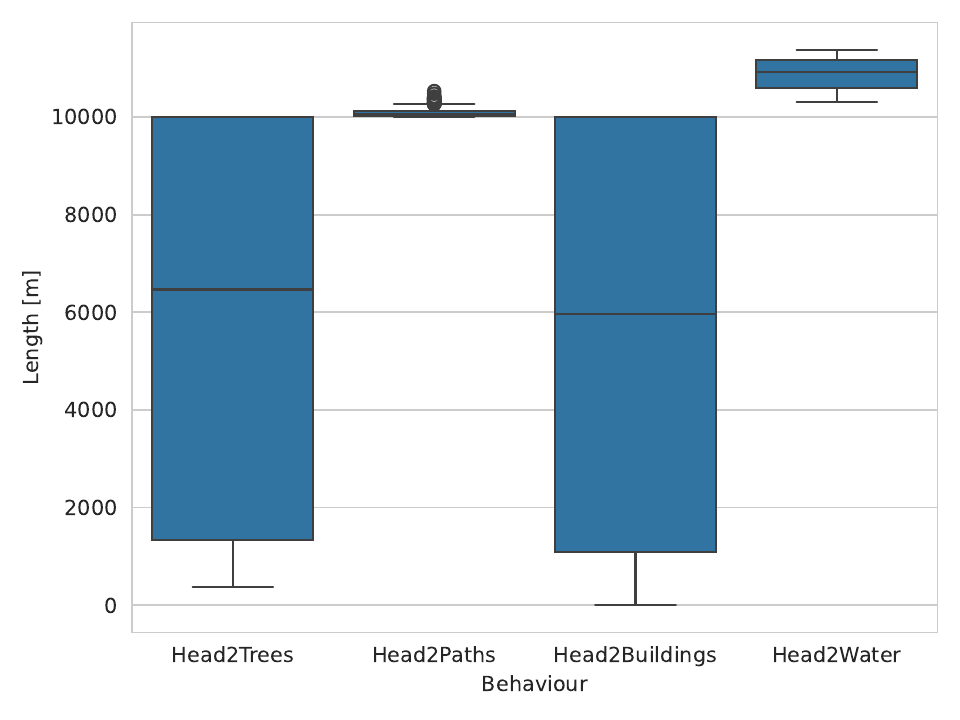}}
	\caption{Analysis of the path lengths by behavior}
	\label{fig:path_lengths}
\end{figure}

As the simulation termination distance was $10,000\si{\m}$, as described in
\refsect{sect:method_sampling}, it is no surprise that over $60\%$ of
paths (\reffig{fig:path_length_distribution}) terminated at this criterion.
However, as seen in \reffig{fig:path_length_boxplot}, it is evident that the two
viewshed behaviors (\headtobuildings and \headtotrees) are terminating early.
\refsect{sect:vector_based_behavior} explains this behavior, and future work
should explore this further.

\begin{figure}[htb!]
	\centering
	\subfloat[Locations found showing the distribution of points in the x-y axis\label{fig:location_with_x_y_dists}]{\includegraphics[width=0.9\linewidth]{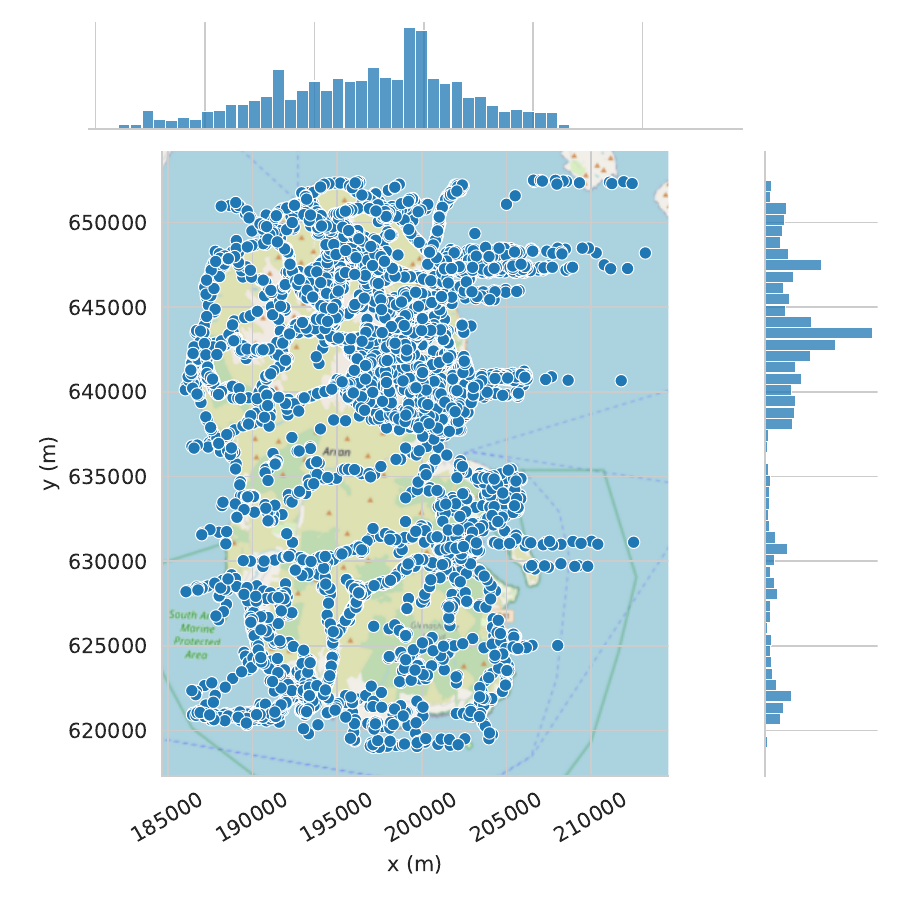}}
	\hfill
	\subfloat[Logarithmic PDM of all PLS locations showing more activity in the north than the south\label{fig:log_heatmap}]{\includegraphics[width=0.9\linewidth]{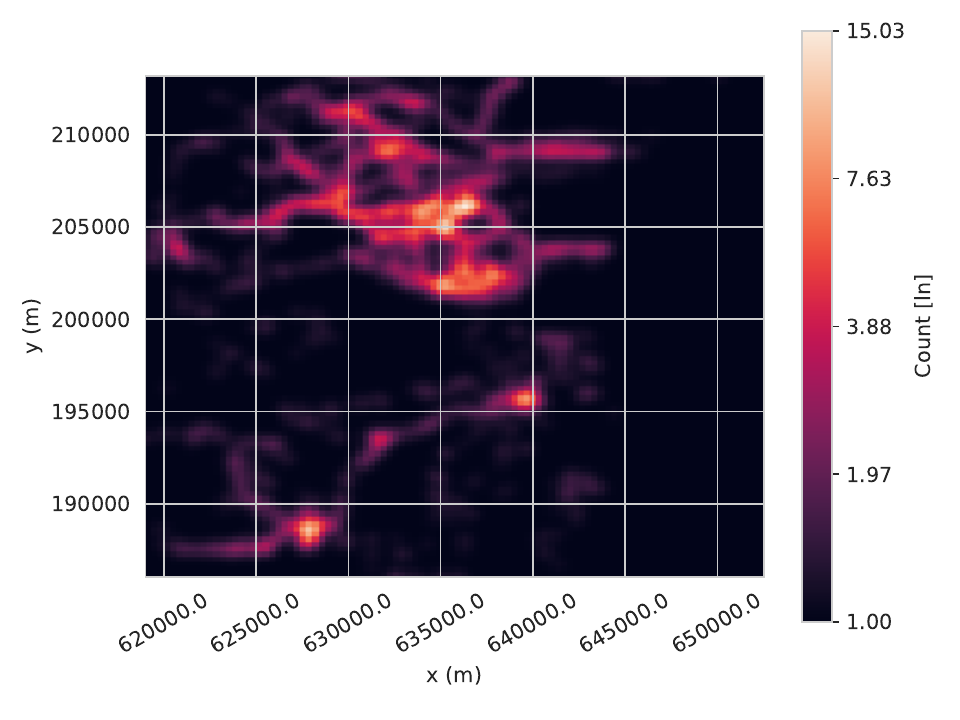}}
	\caption{Paths traversed by agents using the different behaviors with $\sigma_{xx} = \sigma_{yy}=10,000,\rho=0$}
	\label{fig:sampled_routes}
\end{figure}

\reffig{fig:sampled_routes} gives the first glimpse at the generated PDM.
However, these are the aggregated results for multiple PLS locations and as such
would look different for an individual PDM.
Nonetheless, it reflects the sampled PLS by having more locations found in the
north of the island with a slanted band of low probability. Insight like this
would influence the decision-making during a search mission by not looking in that area for
example.

\subsection{Sampling Effectiveness}
\label{sect:results_sampling_effectiveness}

\begin{figure}[htb!]
	\centering
	\includegraphics[width=0.9\linewidth]{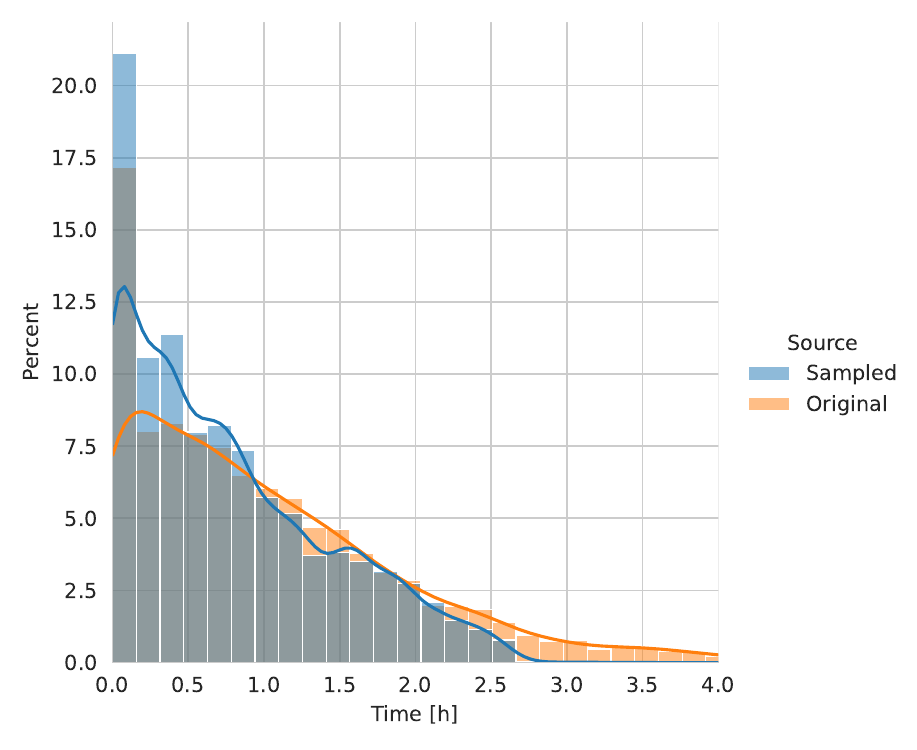}
	\caption{Time distribution of the time used for the sampled points (blue) and the original distribution (orange) showing the filtering in action }
	\label{fig:points_time_sampled_vs_original_distribution}
\end{figure}

The shortened paths from early termination of behaviors result in samples not being taken at distances longer than the path in question.
\reffig{fig:points_time_sampled_vs_original_distribution}
shows the times at which a point was sampled is skewed heavily to the left,
with the original distribution more spread out along the time axis.
The sampled points have a mean time of 
$0.75\si\hour$
and a standard deviation of
$0.66$,
conversely,
the original distribution has a mean of 
$1.06\si\hour$
with a standard deviation of $1.01$. 

This difference in means shows that the upper limit of time traveled correlates to the total time an LP might be missing,
with the distribution becoming closer to matching the historical data as $D_{max}\rightarrow\infty$.
Therefore, $D_{max}$ can be further used to inform the generated PDM for the scenario at hand.

\subsection{Location Found Land Cover Categories}
\label{sect:results_location_found}

\begin{figure}[htb!]
	\centering
	\includegraphics[width=0.9\linewidth]{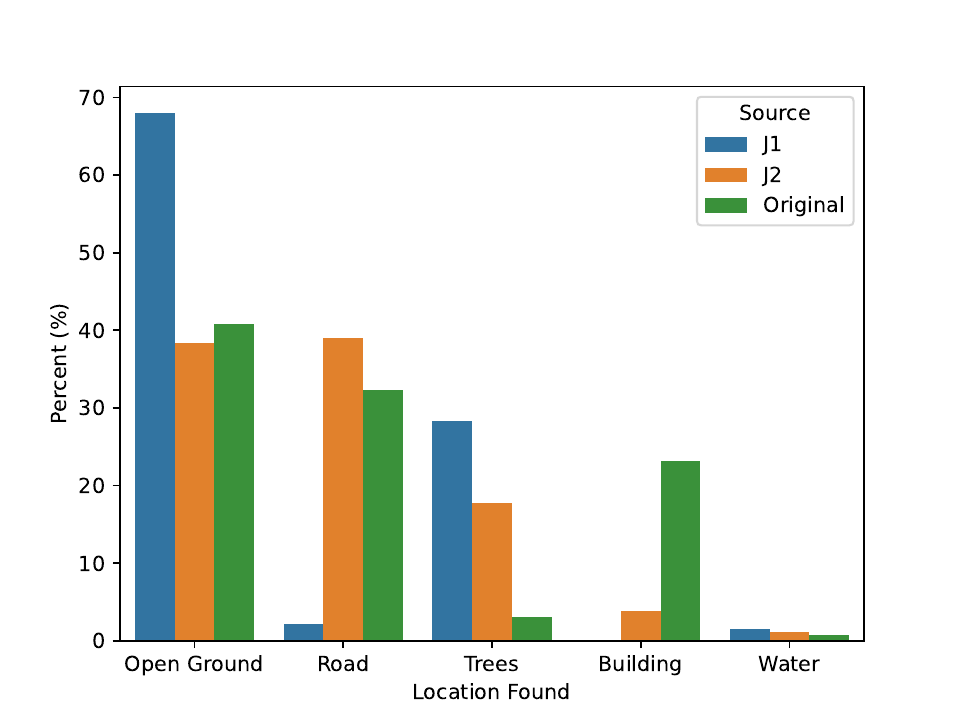}
	\caption{Results from this paper (J2), a previous implementation (J1)\cite{ewers_gis_2023}, and the original data\cite{perkins_missing_2011}}
	\label{fig:locations_found_classified}
\end{figure}

The resultant sampled locations of the agents are the most important metric to
gauge the effectiveness of this model. 
This will be compared to the original data\cite{perkins_missing_2011}, 
and J1, a previous version of the algorithm. 

As can be seen from \reffig{fig:locations_found_classified}, an LP was found in
the \tech{open ground} land cover category $38.36\%$ of the time. This is only a
$2.44\%$ difference to the original data (compared to a $27.22\%$ difference for J1), empirically proving
\refhyp{hyp:open_ground_find_location}. Therefore, LPs end up naturally in the
open-ground land cover category without having a dedicated behavior.

The next land cover category is \tech{road}, at which $38.95\%$ of LPs were found.
Similarly to \tech{open ground}, this is only a difference of $6.65\%$ to
the original data whilst J1 has a $30.20\%$ difference.
Such a large improvement in the \tech{road} land cover category is a result of an improved
\headtopaths behavior algorithm from \refsect{sect:head2roads}. However, the
viewshed-based algorithms, \headtotrees, and \headtobuildings have improved but are
still substantially off the mark compared to the source data. This result may
occur by the way that the sampling is handled.
\reffig{fig:locations_found_classified} shows that even though
$46.20\%$ of paths are due to the \headtobuildings behavior, only $33.07\%$ of
sampled points were a result of the aforementioned paths. This is due to
\headtobuildings having a low mean length of $5.62\si{\kilo\meter}$ compared to the
largest of $11.04\si{\kilo\meter}$ as seen in \reffig{fig:path_length_boxplot}.

The last land cover category, \tech{water}, remains close to the source data, as was the
result for J1, showing that the strategy of following the
water vector map remains valid.

Overall, when using \refeqn{eqn:symmetric-kl}, the algorithm proposed in this paper
had a score of $61.56$ whilst J1 has a score of $306.02$. 
This is a substantial improvement when compared to a
randomly generated distribution with a score of $159.16$ (with $n=1\times10^7$). 
This shows a large improvement toward matching the source data, as is the target.

\section{Conclusion}
\label{sect:conclusion}

This study explored the psychological profile-based PDM generation algorithm J2
which emulates the movement of an LP over a landscape. Through GP-driven
analysis, it is clear that the sparse data problem experienced in SAR can be
overcome leading to the tangible result of saving lives in future operations. 

By characterizing the profile into four distinct behaviors (\headtowater,
\headtotrees,
\headtopaths,
and \headtobuildings) 
the model can be adjusted to match
the land cover category location found description from local datasets. These distinct, simpler,
behaviors are their own model and traverse the landscape as if they were
an LP with a single goal in mind. 
The data shows that running each behavior a percentage amount of times would produce a distribution of land cover category location found descriptions that match the original dataset. 
However, there were discrepancies with the tree and building categories.
As both of these rely on viewshed-based behaviors, 
it follows that this is the root cause of the discrepancy as further explained in 
\refsect{sect:results_sampling_effectiveness}. 
The introduction of the GP-based design of experiment also
benefited the analysis by generalizing the PLS locations. This is something
that was not done in J1 which further decreased the
effectiveness of the analysis with a $n=2$.

Overall, the method of using the land cover category location found statistics directly from the
source data in J2 is valid with a symmetric Kullback-Leibler (\refeqn{eqn:symmetric-kl}) score of $61.56$ whilst a true random
distribution has a score of $159.16$. This means that per-location training is
not required and that a generalized model has been created. However, further
effort needs to be put into improving the viewshed-based algorithms.

Future work will further explore extending the capabilities of the various
behaviors. Following this, more GIS data will be incorporated to further
solidify the agent's roots in the real world. Replacing behaviors with machine
learning models is also something that will need to be explored. Furthermore,
the usage of PDMs to create UAV search trajectories will be explored\cite{ewers_optimal_2023}.

\section*{Data Availability}
The datasets used and/or analyzed during the current study are available from the corresponding author upon reasonable request.

\bibliography{references}

\end{document}